# AI Framework for Early Diagnosis of Coronary Artery Disease: An Integration of Borderline SMOTE, Autoencoders and Convolutional Neural Networks Approach

Elham Nasarian[1], Danial Sharifrazi[2], Saman Mohsenirad[1], Kwok Tsui[1], Roohallah Alizadehsani[3].

1: Grado Department of Industrial and System Engineering, Virginia Tech,
Blacksburg, VA 24061, USA

2: Department of Computer Engineering, Shiraz Branch, Islamic Azad University,
Shiraz, Iran

3: Institute for Intelligent System Research & Innovation (IISRI), Deakin University,
Geelong, Australia

## Abstract

The accuracy of coronary artery disease (CAD) diagnosis is dependent on a variety of factors, including demographic, symptom, and medical examination, ECG, and echocardiography data, among others. In this context, artificial intelligence (AI) can help clinicians identify high-risk patients early in the diagnostic process, by synthesizing information from multiple factors. To this aim, Machine Learning algorithms are used to classify patients based on their CAD disease risk. In this study, we contribute to this research filed by developing a methodology for balancing and augmenting data for more accurate prediction when the data is imbalanced and the sample size is small. The methodology can be used in a variety of other situations, particularly when data collection is expensive and the sample size is small. The experimental results revealed that the average accuracy of our proposed method for CAD prediction was 95.36, and was higher than random forest (RF), decision tree (DT), support vector machine (SVM), logistic regression (LR), and artificial neural network (ANN).

## 1. Introduction and Related Research

Machine learning (ML) techniques are used to inform human decision-making through data analysis in a wide range of contexts [1]. Particularly when classification of evidence based on multiple factors is required in decision-making, synthesizing information, and determining the correct classification is a challenging problem and ML can be utilized to enhance the classification accuracy [2]. In this regard, medical diagnostics may be viewed as a classification problem in which the clinician observes specific risk factors and identifies whether or not the patient is at risk [3]. The accuracy of this type of classification is crucial since a misdiagnosis might lead to delayed therapy that puts the patient in danger or unnecessary therapies with undesirable side effects [4]. ML techniques then can assist clinicians in analyzing medical records and identifying high-risk patients [5].

In this paper we developed an ML methodology to improve the diagnosis of coronary artery disease (CAD). Previous research indicates that in the United States, CAD is responsible for one death every 36 seconds and resulting in enormous yearly healthcare costs of $5.54 billion [6]. Due to the fact that many people's first experience with CAD is a heart attack [4], it is essential to diagnose CAD early and effectively using risk factors that could lead to its progression. However, designing a data analysis strategy for this purpose has a few obstacles. First, an accurate diagnosis of CAD requires the consideration of various factors, such as demographic, symptom, and medical examination, ECG, and echocardiogram results, among others [7]. Therefore, the classification that takes into account all of these characteristics is a complex problem. Second, collecting medical records on chronic diseases such as CAD is an expensive and time-consuming task that frequently requires tracing patients through time and collecting data on many aspects [8]. Consequently, the sample size for ML modeling is presumably limited. In addition, the classification techniques can provide more accurate predictions if the sample consists of enough observations on all classes. For

instance, if the number of CAD patients in the sample is relatively low, the data analysis is subject to a high misclassification error [9]. The number of articles published per year on CAD detection shows an increasing trend. Figure 1 displays the number of publications using machine learning methods and artificial intelligence from 1991 to 2020 [10].

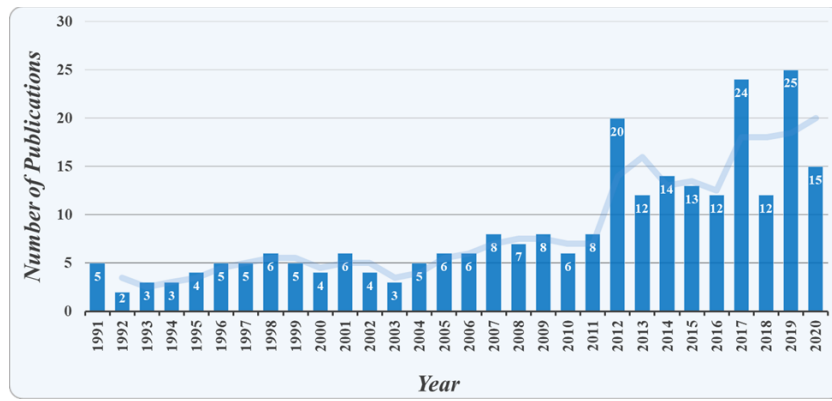

**Figure 1:** The number of publications during 1991-2020

Abdar et al. [17] argue that the diagnosis of CAD is essential in early stages. The paper discussed a framework which combines several traditional machine learning methods (decision trees, artificial neural networks) and deep learning approach for effective diagnosis of CAD. Their findings show that the approach provides 94.66% and 98.60% accuracy for CAD prediction in the Z-Alizadeh Sani and Cleveland CAD datasets. Aouabed et al. [11] used a novel hybrid ML model for CAD detection which it was combined different classical ML algorithms with ensemble learning including DT, SVM, ANN. It was used four different kernel functions (linear, polynomial, radial basis and sigmoid). Authors used the model to analyze the Cleveland CAD dataset. Tama et al. [12] emphasized that owing to the fact that a heart attack occurs without any apparent symptoms, an intelligent detection method is fruitful. Authors proposed a novel CAD prediction model based on a ML. Ghiasi et al. [13] introduced DT called classification and regression tree (CART) and the results are compared with other models in the literature (e.g., Sequential Minimal Optimization (SMO), Naïve Bayes (NB), artificial neural network (ANN), Bagging, and genetic algorithm (GA). Authors contended that CART can outperform some of existing models using Z-Alizadeh Sani dataset.

In a more recent study, Velusamy and Ramasamy [14] remarks that detection of CAD is essential for the medication of patients. The results of base classifiers are combined using ensemble voting technique based on average-voting (AVEn), majority-voting (MVEn), and weighted-average voting (WAVEn) for prediction of CAD. Zomorodi-moghadam et al. [15] also discussed the importance of heart disease risk factors that can increase the probability of CAD. Nowadays, many computer-aided approaches have been used for the prediction and diagnosis of diseases. The study is based on the real-world CAD dataset and aims at the detection of CAD using a hybrid binary-real PSO algorithm.

The main contributions of our paper are two-fold:
(1) The extension of Z-Alizadeh Sani dataset for CAD prediction has been collected in Shaheed Rajaei Cardiovascular, Medical and Research Center. To the best of our knowledge, this dataset is one of the most accurate CAD datasets with stenosis in individual LAD, LCX, and RCA. This dataset contains the records of 303 patients, which 59 features.
(2) This paper has two contributions. First, we combine several ML algorithms to handle the problem of imbalanced data and data augmentation. Our methodology enables the accurate use of ML for CAD diagnostics even when the data at hand is small. Second, we determine the most relevant risk factors that empirically contribute to CAD.

## 2. Dataset

In this research we have used the extension of Z-Alizadeh Sani dataset [10]. This dataset included 303 random patients records with 216 coronary artery disease cases and 87 visitors without CAD, and was collected from Shaheed Rajaei Cardiovascular, Medical and Research Center. This dataset has 59 features for each patient with four categories: (1) demographic information and past medical history; (2) symptomatology and examination findings; (3) laboratory and echocardiographic findings; and (4) electrocardiographic findings (ECG). Moreover, they have used the result of coronary angiography, and if someone had stenosis narrower than 50% of lumen diameter in one of each coronary arteries (RCA, LCX, and LAD), considered as a CAD patient, and otherwise as normal. The target feature in the current study is the binary classification of CAD or normal (yes/no) in patients. Some features are described in **Figure 2** [18].

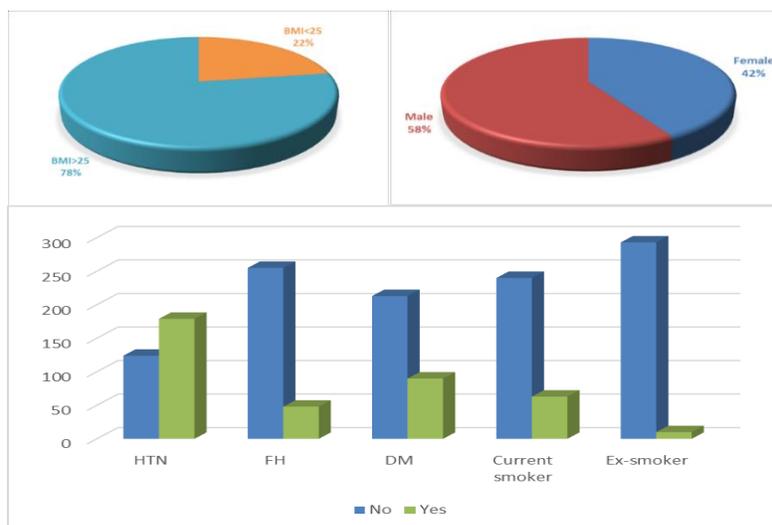

**Figure 2.** Gender, BMI, DM, FH, HTN, and Current Smoker distribution in dataset.

## 3.  Proposed Methodology

In the proposed method, several effective steps are used to improve the diagnosis process of CAD. The proposed dataset is a relatively complex dataset for classification. This method can create an effective classification process on challenging datasets.

### 3.1. Data cleaning and determine whether the dataset is balanced

The dataset contains two classes, that is why binary classification is used for classification. There are 303 samples and 58 predictors in the dataset. By checking the dataset, it is clear that the "Exertional CP" column has the same number in all samples. So, the first step in the proposed method is data cleaning. At this stage, the "Exertional CP" column is removed from the dataset and the number of predictors is reduced to 57.

### 3.2. Using the Borderline SMOTE method to balance the dataset

With a more detailed examination of the dataset, it became clear that we are dealing with an imbalance dataset. That is, the number of samples of two classes is not equal. For a successful classification, the number of samples of both classes should be equal. Because if the number of samples of the classes is not equal, we may face the risk of increasing the accuracy and decreasing the Area Under Curve (AUC) metric, and this high accuracy will not be a real accuracy. To balance the dataset, the Borderline SMOTE method is used in the proposed method. In the beginning, the number of samples of the classes were 87 and 216 samples (303 samples in total). After balancing the dataset, the number of samples of the classes reached 216 and 216 (432 samples in total).

### 3.3. Data augmentation through Autoencoder

In the next step, the number of samples is not enough for an efficient classification. If the number of data is small, the probability of overfitting is very high. In fact, the machine learning model cannot learn the data correctly and only memorizes it. To overcome this problem, a common method is data augmentation. For data augmentation, the Autoencoder method is used in the proposed method. The proposed Autoencoder is a deep neural network that includes three input, one output and one hidden layer. The number of input and output layer neurons is 57 (the number of dataset features). 32 neurons are used in the hidden layer too. All data of dataset is passed through Autoencoder layers and in the output layer, for each data we will have reconstructed data that can be used as new data. So, the final dataset is created.

### 3.4. Splitting dataset into test and train parts through the K-Fold Cross Validation method

After data augmentation, the number of data reaches 826, which is a suitable number for classification. In the classification stage, the final dataset is divided into 10 parts through the K-Fold Cross Validation method to ensure sufficient results. To clarify more, the classification method is performed 10 times. In each run, 10% of the data is selected as test data and 90% of the dataset is selected as training data. The final accuracy of the proposed method is the average accuracy of 10 executions of this method.

### 3.5. Implementing different ML/DL methods and selecting the best method

In the classification stage, six different methods called Decision Tree (DT), Random Forest (RF), Support Vector Machine (SVM), Logistic Regression (LR), Artificial Neural Network (ANN), and Convolutional Neural Network (CNN) are used; it is clear that the best method for classification is the CNN method. The proposed CNN consists of nine main layers. Four convolutional layers and five fully-connected (Dense) layers. In order to avoid overfitting, after each the first four fully-connected layers, a "Dropout" layer is also used and the last layer is considered as the output layer. The details of the proposed CNN parameters can be seen in **Table 1**. In this research Sci-kit Learn library for machine learning methods and Keras library with TensorFlow backend were used for deep learning programming as well. Furthermore, the NVIDIA GeForce GTX 950 GPU was applied for deep learning implementation.

**Table 1:** CNN Hyperparameters.

| *Hyper Parameters* | *Values* | *Hyper Parameters* | *Values* | *Hyper Parameters* | *Values* |
|---|---|---|---|---|---|
| Input dimension | 57×1 | Kernel Regularizes | L2=0.2 | Loss function | Binary cross entropy |
| Number of convolutional layers | 4 | Activation function for hidden layers | ReLU | Function class | 1, 2, 3, 4 |
| Number of fully connected layers | 5 | Activation function of the last layer | Sigmoid | Number of neurons of fully connected layers | 256, 128, 64, 32, 2 |
| Number of filters for each convolutional layer | 256, 256, 256, 256 | Optimizer | Adam | Dropout probability | 0.5 |
| Strides size | 1 | Learning rate | 0.001 | Number of epochs | 100 |
|  |  | Batch size | 256 |  |  |

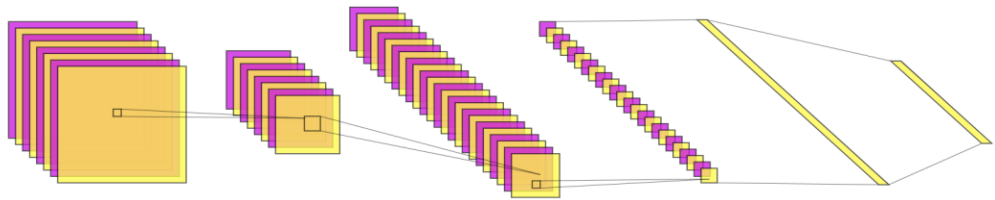

**Figure 3.** An example of CNN Architecture

## 4. Results

As mentioned in methodology part, for preventing of any biases and getting reliable results, the final dataset (after balance and augmentation) was divided into 10 folds with the K-Fold Cross Validation method for all train and test parts, **Table 2** shows the average of these 10 folds for all methods. Also, several metrics were used: Recall, Precision, F1-Score, Accuracy, and ROC AUC for indicating that our proposed method with balance dataset could have improvement in classification and prediction.

**Table 2.** Average performances after 10-fold cross validation for proposed method and ML models

| *Prediction Algorithms* | *Recall* | *Precision* | *F1 Score* | *Accuracy* | *ROC AUC* |
|---|---|---|---|---|---|
| Decision Tree (DT) | 87.10 | 87.75 | 87.30 | 89.44 | 87.15 |
| Random Forest (RF) | 93.85 | 95.6 | 85.00 | 94.77 | 84.94 |
| Support Vector Machine (SVM) | 92.45 | 94.60 | 93.40 | 94.66 | 93.51 |
| Logistic Regression (LR) | 88.60 | 90.00 | 88.70 | 90.25 | 88.17 |
| Artificial Neural Network (ANN) | 91.95 | 90.35 | 91.05 | 91.43 | 89.85 |
| **Proposed Method** | **95.00** | **94.80** | **95.05** | **95.36** | **95.06** |

According to **Table 2**, it can be observed our proposed method had higher accuracy compared to DT, RF, SVM, LR, and ANN methods. **Figure 4**, shows the average of our proposed method, and the accuracy of this method in 10 folds.

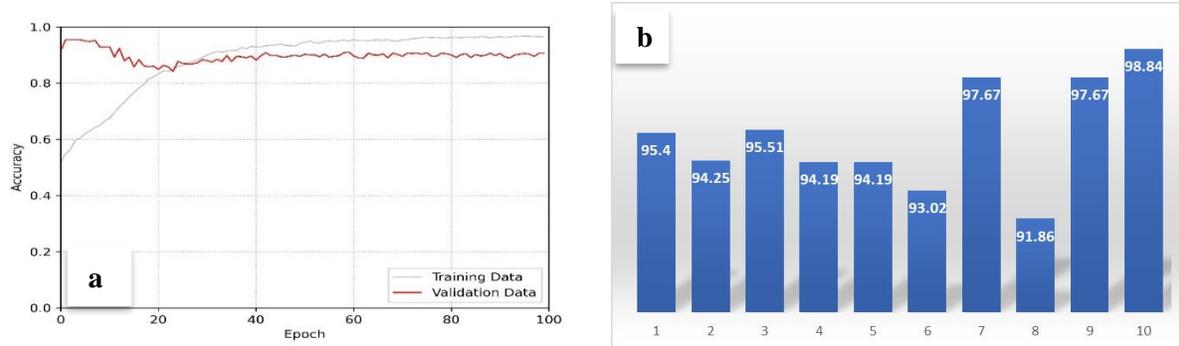

**Figure 4. a)** Average accuracy of CNN training process during 10 folds, and **b)** Obtained accuracy of proposed method for 10 folds.

## 5. Conclusions

This research paper presented an integration of borderline SMOTE, autoencoder (AE), and convolutional neural network (CNN) for small and imbalance dataset for CAD detection in early stages based on medical records. **Figure 5** indicates our proposed method achieved better performance compare to classical predictive analysis. In the next steps, we will add new features, new targets for predicting each artery stannic separately, and also improve our framework for getting higher accuracy.

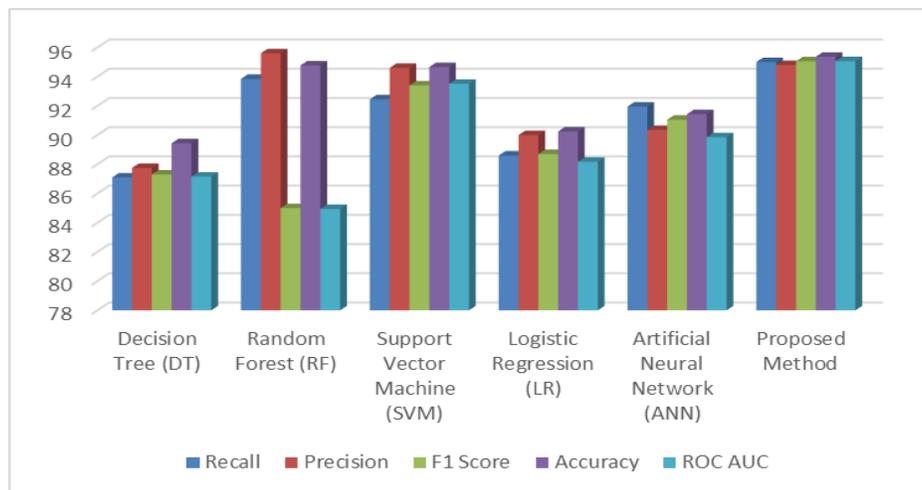

**Figure 5.** Comparison between our proposed method and classical ML methods.